\documentclass[letterpaper, 10 pt, journal, twoside]{IEEEtran} 

\IEEEoverridecommandlockouts   

\usepackage{mwe}
\usepackage{graphics} 
\usepackage{graphicx}
\usepackage{times} 
\usepackage{amsmath} 
\usepackage{amssymb}  
\usepackage{multirow}
\usepackage{lipsum}
\usepackage{color}
\usepackage{subcaption}
\usepackage[noabbrev,capitalize]{cleveref}
\usepackage{mathtools}
\usepackage{booktabs}
\usepackage{ulem}
\usepackage{makecell}

\let\llncssubparagraph\subparagraph
\let\subparagraph\paragraph
\usepackage[compact]{titlesec}
\let\subparagraph\llncssubparagraph

\begin{document}
\title{\LARGE \bf
Label-Efficient Point Cloud Segmentation\\ with Active Learning
}

\author{
Johannes Meyer$^{1*}$, 
Jasper Hoffmann$^{1*}$, 
Felix Schulz$^{1}$, 
Dominik Merkle$^{2,3}$, 
Daniel Buescher$^{1}$,\\ 
Alexander Reiterer$^{2,3}$, 
Joschka Boedecker$^{1}$, 
Wolfram Burgard$^{4}$
\thanks{$^{*}$ Equal contribution.}%
\thanks{$^{1}$ Department of Computer Science, University of Freiburg, Germany}
\thanks{$^{2}$ Fraunhofer IPM, Freiburg, Germany}
\thanks{$^{3}$ Institute for Sustainable Systems Engineering INATECH, University of Freiburg, Germany}
\thanks{$^{4}$  Department of Computer Science and
Artificial Intelligence, University of Technology Nuremberg, Germany.}

\thanks{This research was funded by the German Federal Ministry for the Environment, Nature Conservation and Nuclear Safety (BMU) on the basis of a resolution of the German Bundestag as part of the ‘KI-Leuchtturm’ project ‘Intelligence for Cities’ (I4C).}
}

\maketitle

\begin{abstract}
Semantic segmentation of 3D point cloud data often comes with high annotation costs.
Active learning automates the process of selecting which data to annotate, reducing the total amount of annotation needed to achieve satisfactory performance.
Recent approaches to active learning for 3D point clouds are often based on sophisticated heuristics for both, splitting point clouds into annotatable regions and selecting the most beneficial for further neural network training.
In this work, we propose a novel and easy-to-implement strategy to separate the point cloud into annotatable regions. In our approach, we utilize a 2D grid to subdivide the point cloud into columns. To identify the next data to be annotated, we employ a network ensemble to estimate the uncertainty in the network output.
We evaluate our method on the S3DIS dataset, the Toronto-3D dataset, and a large-scale urban 3D point cloud of the city of Freiburg, which we labeled in parts manually.
The extensive evaluation shows that our method yields performance on par with, or even better than, complex state-of-the-art methods on all datasets.
Furthermore, we provide results suggesting that in the context of point clouds the annotated area can be a more meaningful measure for active learning algorithms than the number of annotated points. 
\end{abstract}


\section{Introduction}

Semantic point cloud segmentation is pivotal for many applications including robotics, urban planning, and environmental monitoring. The semantic segmentation of urban point cloud data is particularly important as a basis for wind, water, and heat simulations~\cite{BriegelModelling}. This can aid in identifying vulnerable areas within cities, thereby enhancing their resilience to climate change.
The simulations require semantic information to differentiate between various surface types, such as sealed or open surfaces, which can impact water seepage.
The distinction between fir and leaf trees due to the different capabilities of water storage and leaf fall is important to correctly simulate heavy rain and wind events, and simulate the heat load in cities at different seasons.
The size and diversity of cities require a substantial amount of labeled data to sufficiently train state-of-the-art neural networks. 
Unfortunately, the annotation process for urban 3D point cloud data is especially costly~\cite{unal2022scribblesupervised,Merkle2022}.
In practice, to precisely segment an object in 3D requires drawing many different 2D polygons from various perspectives. 

\begin{figure}[t]
\centering
\def\svgwidth{0.9\columnwidth}
\small
\begingroup%
  \makeatletter%
  \providecommand\color[2][]{%
    \errmessage{(Inkscape) Color is used for the text in Inkscape, but the package 'color.sty' is not loaded}%
    \renewcommand\color[2][]{}%
  }%
  \providecommand\transparent[1]{%
    \errmessage{(Inkscape) Transparency is used (non-zero) for the text in Inkscape, but the package 'transparent.sty' is not loaded}%
    \renewcommand\transparent[1]{}%
  }%
  \providecommand\rotatebox[2]{#2}%
  \newcommand*\fsize{\dimexpr\f@size pt\relax}%
  \newcommand*\lineheight[1]{\fontsize{\fsize}{#1\fsize}\selectfont}%
  \ifx\svgwidth\undefined%
    \setlength{\unitlength}{178.86061478bp}%
    \ifx\svgscale\undefined%
      \relax%
    \else%
      \setlength{\unitlength}{\unitlength * \real{\svgscale}}%
    \fi%
  \else%
    \setlength{\unitlength}{\svgwidth}%
  \fi%
  \global\let\svgwidth\undefined%
  \global\let\svgscale\undefined%
  \makeatother%
  \begin{picture}(1,0.69752412)%
    \lineheight{1}%
    \setlength\tabcolsep{0pt}%
    \put(0,0){\includegraphics[width=\unitlength,page=1]{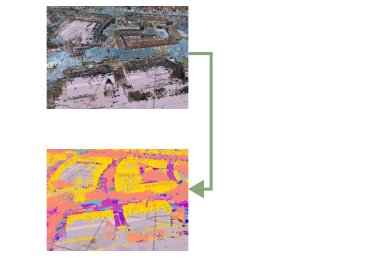}}%
    \put(0.03262747,0.42678264){\rotatebox{90}{\makebox(0,0)[lt]{\lineheight{1.25}\smash{\begin{tabular}[t]{c}Colorized urban\\point cloud\end{tabular}}}}}%
    \put(0.03262747,0.05219346){\rotatebox{90}{\makebox(0,0)[lt]{\lineheight{1.25}\smash{\begin{tabular}[t]{c}Semantic urban\\point cloud\end{tabular}}}}}%
    \put(0,0){\includegraphics[width=\unitlength,page=2]{coverpage.pdf}}%
    \put(0.50725476,0.40567698){\makebox(0,0)[t]{\lineheight{1.25}\smash{\begin{tabular}[t]{c}Label-efficient\\segmentation with\\active learning (AL)\end{tabular}}}}%
    \put(0,0){\includegraphics[width=\unitlength,page=3]{coverpage.pdf}}%
    \put(0.70281676,0.09040607){\makebox(0,0)[lt]{\lineheight{1.25}\smash{\begin{tabular}[t]{c}Samples by\\AL algorithm\end{tabular}}}}%
    \put(0,0){\includegraphics[width=\unitlength,page=4]{coverpage.pdf}}%
  \end{picture}%
\endgroup%
\caption{
The goal of this work is to reduce the annotation cost of semantic segmentation for unlabeled urban point clouds. 
By simplifying existing methods, we aim to reduce the entry barrier to apply active learning for point clouds.}
\label{fig:coverpage}
\end{figure}

One possible solution is active learning (AL), illustrated in \cref{fig:coverpage}. 
AL can drastically reduce labeling costs by only ``requesting'' the labels for the most informative unlabeled samples.
In practice, the AL algorithm starts to train a model with only a small labelled portion of the data.
After each training cycle, the AL algorithm aims to find the most informative part of the data and obtains an annotation from a human or oracle. This process is repeated until the desired performance is reached.

Importantly, the problem of 3D AL is not just about finding individual points in the point cloud, but also finding regions that can be efficiently annotated by a human.
Thus, there are two major challenges in AL for 3D, namely
\emph{region separation}: splitting the point cloud into annotatable candidate regions, and
\emph{region selection}: selecting the most beneficial regions to be annotated.

\begin{figure*}[]
    \centering
    \def\svgwidth{2.\columnwidth}
    \footnotesize
    \resizebox{1.4\columnwidth}{!}{%
    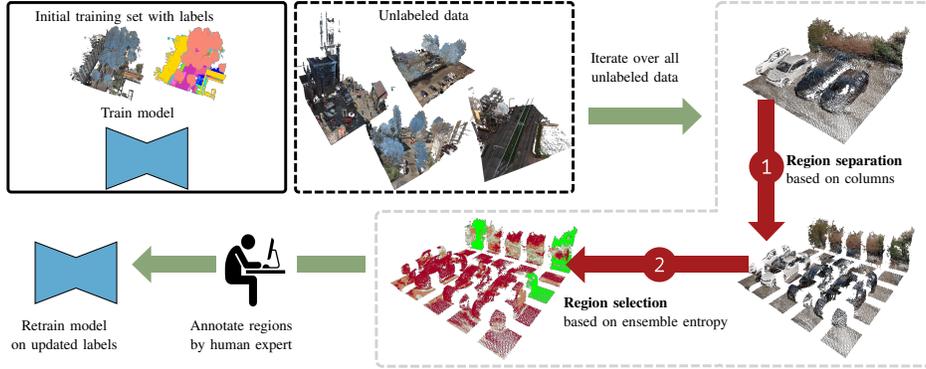
    }
    \caption{Our proposed active learning pipeline.
    The initial dataset consists of unlabeled and labeled parts.
    The AL algorithm first separates the point cloud into columns and then selects regions with the highest ensemble entropy.
    These are presented to a human expert for extending the labeled dataset.
    We iteratively repeat the procedure.}
    \label{fig:method}
\end{figure*}

Previous works in 3D AL handled the region separation and region selection step by incorporating sophisticated heuristics.
However, these approaches often require cumbersome pre-processing steps and lead to proposals that can be harder to annotate.
The approach proposed in this paper separates the point cloud into easy to implement spatial columns and bootstraps the AL pipeline by requiring fewer pre-processing steps.
Our proposed AL cycle is illustrated in \cref{fig:method}.

Our contributions can be summarized as follows:
\begin{enumerate}
    \item For region separation, we find that our straightforward approach in finding annotable regions is competitive with respect to state-of-the-art AL methods.
    \item For region selection, we analyze different common AL metrics based on ensembles, showing better or at least comparable performance when compared with specialized metrics used in current state-of-the-art point cloud segmentation works.
    \item We show that the number of labeled points can be a misleading measure and propose an alternative metric based on the annotated area.
    \item We show the applicability of our approach on a large-scale urban 3D dataset of the city of Freiburg.
\end{enumerate}


\section{Related Work}
To reduce labeling costs of large point cloud datasets, various approaches have
been proposed like using scribble annotation to sparsely annotate data~\cite{unal2022scribblesupervised} or pre-trained models from the 2D image domain \cite{sametYouNeverGet2023}.
For the related work, we will focus on active learning especially in the context of point cloud segmentation.


\subsection{Active Learning}

In the field of AL, a lot of work focuses on the selection of the next batch of data to query a human annotator or an oracle.
Most approaches in the field of AL fall into one of three categories~\cite{ren2021survey}. Firstly, approaches that try to maximize the diversity within one batch of queried data. Secondly, approaches that try to select data points that the current model is uncertain about. And thirdly, those which try to estimate which data points will result in the biggest changes to the current learned model. 

However, the literature shows that only following one approach of AL is problematic for 2D data~\cite{Wang2017} as well as for 3D data~\cite{wu_redal_2021}.
Only selecting samples based on the diversity strategy is prone to yield low-information samples which are only selected for diversity reasons and do not contribute valuable information to the training.
On the other hand, approaches based on the uncertainty of the network often rely on the softmax probabilities of the network which are known to be overconfident \cite{guoCalibrationModernNeural2017}.
For that reason, in deep AL often hybrid querying strategies based on the network uncertainty as well as  diversity estimates, derived from heuristics, are used \cite{ren2021survey}.

In our work, we investigate whether a pure ensemble uncertainty-based approach can outperform current state-of-the-art hybrid approaches specialized for point clouds.

\subsection{Active Learning for Point Cloud Segmentation}\label{subsec:al_pcl}
Several works propose using AL in the context of 3D point cloud segmentation, mostly using a hybrid AL approach, considering uncertainty and diversity for selecting the next batch of data.
Importantly, they use specialized heuristics and require additional pre-processing steps.
In~\cite{wu_redal_2021}, the authors propose the \textit{Region-based and Diversity-aware Active Learning for Point Cloud Semantic Segmentation} (ReDAL) method.
First, the point cloud is separated into regions, called supervoxels, by using an unsupervised segmentation method.
For selecting the supervoxel, ReDAL considers heuristics based on color discontinuities between each point and its $k$-neighbors as well as structural complexity, based on the surface variation.
Combined with the softmax entropy of the current semantic segmentation model a region information score is obtained.
To ensure diversity, the backbone model features for each point in each supervoxel are averaged, and a $k$-means clustering across all supervoxels is computed.
The information score for the regions is lowered for each region belonging to the same cluster that has a higher information score, ensuring that the information score considers diversity.

A similar hybrid strategy was proposed by Shao \emph{et al.}~\cite{shao_active_2022} and is called \emph{Active Learning for point
cloud semantic segmentation via Spatial-Structural Diversity Reasoning} (SSDR-AL). 
Similar to ReDAL in SSDR-AL, the point cloud is separated into a set of superpixels, which we call supervoxel for simplicity.
The region selection policy operates in two steps: In the first step, proposal supervoxels are sampled based on the average network uncertainty for each supervoxel, combined with a weighting, that based on the currently predicted classes, boosts the probability of underrepresented classes.
In the second step, each supervoxel is projected into a so-called diversity space.
The projection into the diversity space is done by averaging the features predicted by the backbone model for each point contained in the supervoxel and its surrounding neighboring supervoxels.
To select from the regions sampled in the first step, we use a furthest-point sampling in the diversity space.

Samet \emph{et al.}~\cite{sametYouNeverGet2023} find that the initial annotated set of regions needed for warm-starting the AL can have a significant influence on the final performance.
Generating images from different views of the 3D scene, they use features from a pre-trained DINO model \cite{caronEmergingPropertiesSelfSupervised2021}
to generate a diverse initial data set, significantly improving the performance of several AL methods.
Xu \emph{et al.}~\cite{xuHierarchicalPointbasedActive2023} presented a successful combination of AL with self-supervised learning, in which the annotator was queried to annotate individual points, instead of regions. However, as we argue in \cref{sec:measuring_effort} annotating a lot of individual points still can come with a high labeling effort.
A sparse point cloud is able to cover a large area of the unlabeled point cloud, even if the percentage of queried points is low, increasing the workload for the human annotator.


\section{Approach}

The two major challenges in AL for point cloud segmentation are region separation, where we separate the point cloud into proposal regions that can be efficiently annotated, and region selection, where we want to find regions with the highest impact on segmentation performance.

\subsection{Region Separation}

In the following, we discuss our region separation mechanism to split the point cloud into 3D columns and also discuss previous state-of-the-art region separation techniques based on supervoxels.

\subsubsection{Columns}
This work employs a straightforward method of dividing the point cloud into easily and efficiently annotated regions, specifically columns.
We divide the point cloud into spatial columns of a given grid resolution~$r$ using a 2D grid on the XY-plane.
This ensures that each query of the AL algorithm does not overburden the human annotator, as the number of points that can be queried is limited by the size of the column.
Additionally, each column can be described using straightforward $x$-$y$-coordinates, without the need for to store clusters of point indices.

\subsubsection{Supervoxels}
Region-based AL often uses unsupervised segmentation methods to separate the point cloud into coherent regions, called supervoxels.
ReDAL~\cite{wu_redal_2021} uses the Voxel Cloud Connectivity Segmentation (VCCS) method~\cite{papon2013voxel}, which aims to produce over-segmentation masks that are fully consistent with the spatial geometry within each mask.
Alternatives for generating such supervoxels are DBSCAN~\cite{ester1996density} or HDBSCAN~\cite{campello2013density}. It is also possible to combine such techniques with a prior ground segmentation as discussed in previous work~\cite{nunesSegContrast3DPoint2022}.
However, such approaches sometimes struggle to find coherent supervoxels that can be easily annotated on real-world data.
Similarly, in a more recent work~\cite{shao_active_2022} the point cloud is also split into supervoxels by using an unsupervised segmentation method based on a global energy model~\cite{Guinard2017}.


\subsection{Region Selection}
In the following, we discuss different region selection methods.
Firstly, we discuss random region selection, which is an important AL baseline~\cite{ren2021survey},
and previous state-of-the-art region selection techniques that are based on point cloud heuristics.
Secondly, we introduce two well established uncertainty metrics that are based on ensembles, the variation ratio and the ensemble entropy.
Both are agnostic to point cloud segmentation and can lead to significantly improved uncertainty estimates for AL \cite{ren2021survey}.


\subsubsection{Random}
Random selection of regions is often used as a baseline in the AL community. It serves as an indicator of whether the proposed selection metric is better than a completely uninformed method.
Notable, performance estimates of random selection policies often heavily vary between different works, which also has been discussed by Ren \emph{et al.}~\cite{ren2021survey} and Samet \emph{et al.}~\cite{sametYouNeverGet2023}.

\subsubsection{Heuristics-based Approaches}
As discussed in \cref{subsec:al_pcl}, current state-of-the-art methods like ReDAL~\cite{wu_redal_2021} and SSDR-AL~\cite{shao_active_2022} are hybrid approaches that use heuristics, such as surface variation or color gradients, to get a reliable diversity estimation. However, since these metrics are often pre-computed they often only serve as a prior to circumvent the problem of overconfident neural networks \cite{guoCalibrationModernNeural2017}. 
Using deep ensembles and averaging the probabilities like in equation \eqref{eq:ensemble probability},
can significantly improve the calibration~\cite{lakshminarayanan2017simple},
leading to a more meaningful uncertainty metric for AL~\cite{ren2021survey}.


\begin{figure}[t!]
\centering%
\hspace*{-0.03\linewidth}%
\includegraphics[width=0.55\linewidth]{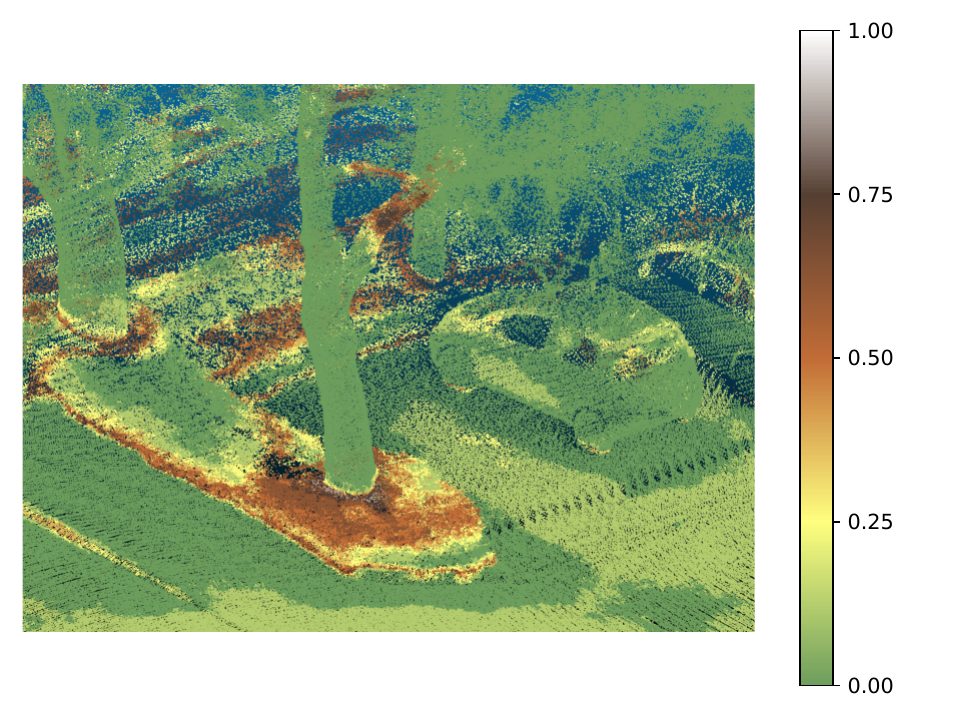}%
\hspace*{-0.12\linewidth}%
\includegraphics[width=0.55\linewidth]{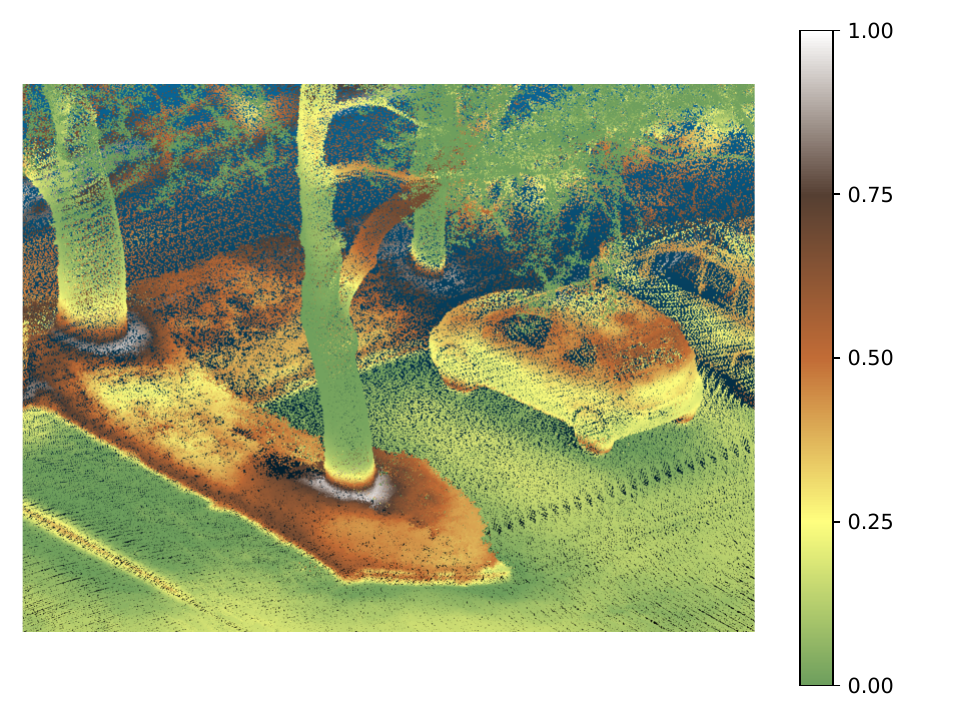}%
\hspace*{-0.05\linewidth}%
\caption{VaR (left) and ensemble entropy (right) for the Freiburg dataset. Green corresponds to small, and white to large uncertainty. This image shows that both uncertainty metrics indicate a high uncertainty in the areas which are lower-vegetation. In contrast, the entropy indicates a higher uncertainty in the upper parts of the car than the VaR.}
\label{fig:var_entropy}
\end{figure}

\subsubsection{Variation ratio (VaR)}
The variation ratio (VaR) is based on deep ensembles, where $N$ different neural networks are initialized with different parameters but are trained on the same dataset, to select regions for annotation.
The variance in the predictions of the ensemble members can be interpreted as the model uncertainty for each point cloud region.
This can be used as guidance for which samples, or regions of the point cloud in our case, should be labeled next~\cite{beluchPowerEnsemblesActive2018}.
For a given point $x$, the VaR measures the fraction of ensemble members diverging from the majority class as
\begin{equation*}
    \textrm{VaR}(x) \coloneqq 1 - \frac{f_m(x)}{N},
\end{equation*}
where $f_m(x)$ is the frequency of the most selected class for the point $x$.
The lower $f_m(x)$ is, the higher the discrepancy in network predictions of the ensemble and the higher VaR will be.
If all members agree on a prediction, the VaR will be $0$.
To derive a VaR estimate for a given region $S$, we define the average VaR as 
\begin{equation}\label{eq:AvgVaR}
    \overline{\textrm{VaR}}(S) \coloneqq \frac{1}{|S|} \sum_{x \in S} \textrm{VaR} (x).
\end{equation}

\subsubsection{Ensemble entropy (Ent)} Another common metric used in AL based on ensembles is the average entropy~\cite{beluchPowerEnsemblesActive2018}.
It is derived by the averaged predicted probabilities of the ensemble members, namely
\begin{equation}\label{eq:ensemble probability}
    \hat{p}\left(y = c\mid x \right) \coloneqq \frac{1}{N} \sum_{n=1}^N p\left(y = c \mid x, \theta_n \right),
\end{equation}
where $p\left(y=c \mid x, \theta_n \right)$ denotes the predicted probability of the $n$th member of the ensemble of predicting class $c$ for an input $x$.
The metric for a single point $x$, is then defined as the entropy of the average prediction
\begin{equation*}
    \textrm{Ent}(x) \coloneqq - \sum_{c=1}^{C} \hat{p}\left(y = c \mid x \right)\; \log( \hat{p}\left(y = c \mid x \right)),
\end{equation*} 
where $C$ is the number of classes.
Similar to \eqref{eq:AvgVaR}, we define then the average entropy for a region  $S$ as
\begin{equation}
    \overline{\textrm{Ent}}(S) \coloneqq \frac{1}{|S|} \sum_{x \in S} \textrm{Ent}(x).
\end{equation}

In \cref{fig:var_entropy}, we show the examples for street-scene computed using the network output for four different random seeds.
For this visualization, we use SPVCNN as a segmentation model.
Both the VaR and the entropy are normalized between 0 and 1.
Both metrics indicate high uncertainty around the edges of the low vegetation areas and for regions in close proximity to other classes.
The entropy metric indicates a high uncertainty around the upper part of the vehicle.

\subsection{Measuring Annotation Effort}
\label{sec:measuring_effort}

The main goal of the region separation and selection pipeline in the active learning scheme is
to provide proposals for efficient annotation by humans.
However, we argue that the expected annotation effort is not well estimated in existing studies:
it is measured in terms of the fraction of Lidar points to be annotated \cite{wu_redal_2021,ren2021survey,xuHierarchicalPointbasedActive2023,shaoActiveLearningPoint2022}.
This measure can be very misleading, in particular when the selected points are sparsely scattered.
In this case a small fraction of points can cover the surfaces of many objects.
Since a human is not efficient in annotating single points,
we argue that the covered surface area needs to be considered to estimate the annotation effort.

\begin{figure}[t!]
    \centering
      \includegraphics[width=0.49\linewidth,trim=10cm 0 10cm 0,clip]{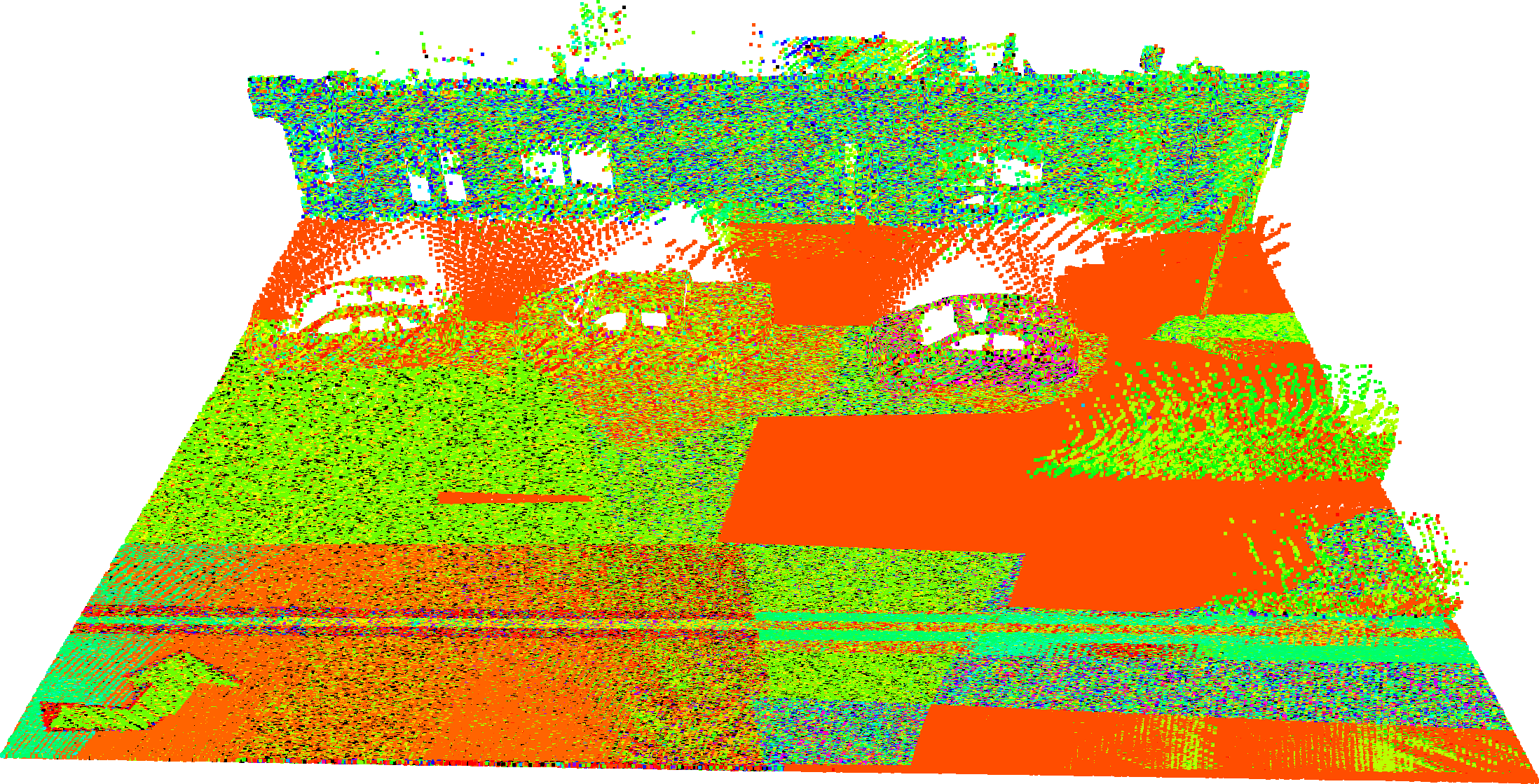}
      \includegraphics[width=0.49\linewidth,trim=10cm 0 10cm 0,clip]{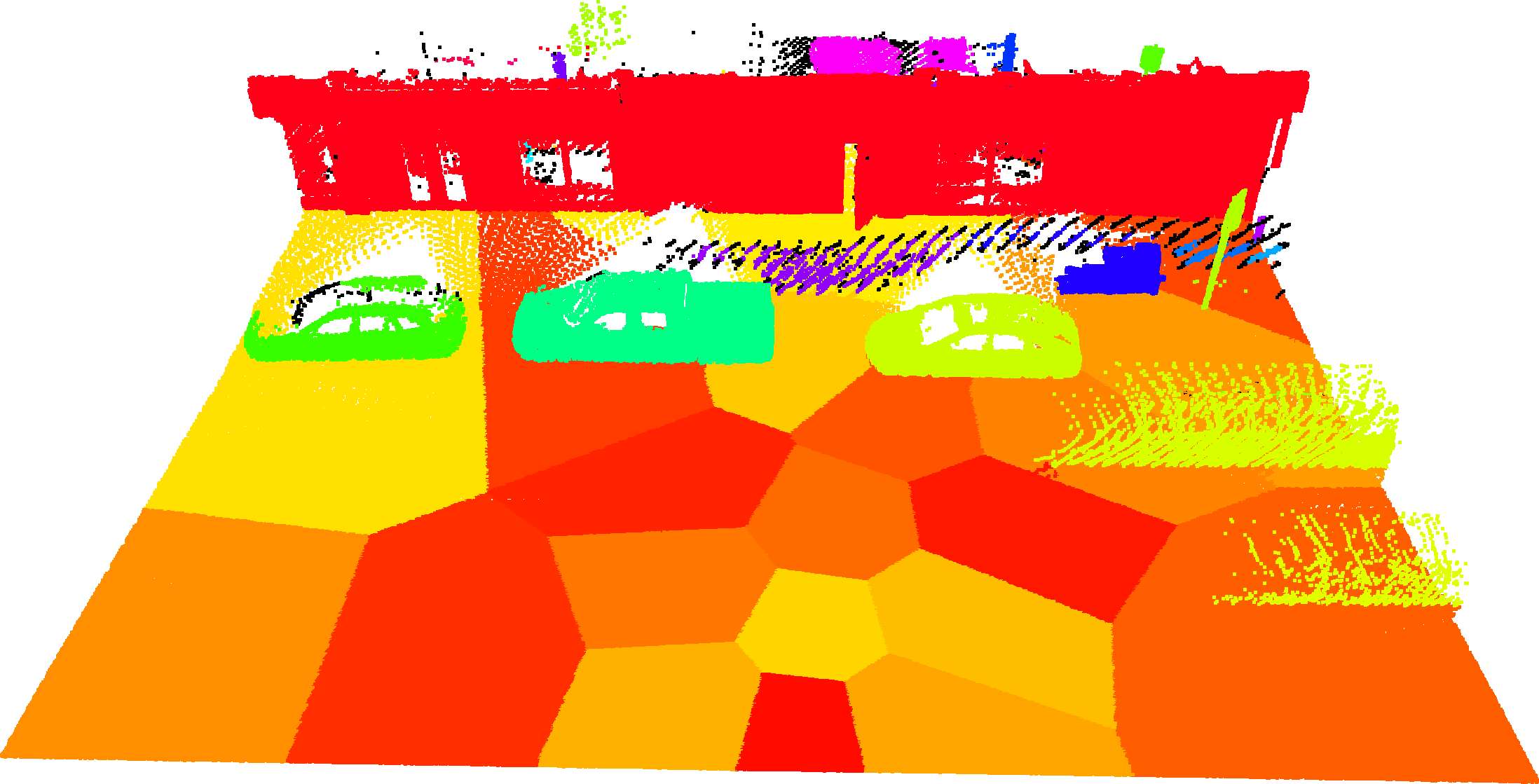}
    \caption{
    Region separation on the Toronto data with VCCS (left) and HDBScan (right). Each set of points (supervoxels) is drawn in a different color. The very noisy representation of the clusters on the left depicts the failure of VCCS on this dataset. In contrast on the right the combination of HDBScan with ground-plane removal gives very concise clusters.
    }
    \label{fig:annotation_cost}
    \vspace{-.5cm}
\end{figure}

The performance of the AL pipeline, when using supervoxels calculated by VCCS, measured as function of the fraction of selected points is very competetive. However, as the illustration in \cref{fig:annotation_cost} shows the supervoxels created by VCCS cover large but sparse areas in the point cloud. In an AL pipeline which utilizes an oracle for label retrieval, such voxels can be efficiently annotated.
However, a human might not be able to efficiently annotate large sparse point clouds. Despite our best efforts, we found that VCCS, which is the clustering method used in ReDAL, was not able to produce reliable clusters.




In order to compare ourself on Toronto-3D, we propose an alternative supervoxel-based region separation mechanism for that dataset.
We segment a ground plane in the data (if available) and cluster the rest with HDBScan.
This was also proposed by Nunes \emph{et al.} in the context of contrastive learning~\cite{nunesSegContrast3DPoint2022}.
Furthermore, we divide the ground plane into several smaller regions by using K-means clustering.
Using this approach we are able to retrieve coherent supervoxels as visualized in \cref{fig:annotation_cost}.

Furthermore, one of our central contributions is the following metric to estimate the annotation effort.
As argued above, we base our metric on the area covered by the selected points.
An exact estimate of this area is not easy to obtain, but we simplify it by using (half of)
the surface $A$ of the cuboid containing the selected points,
where $A = \Delta x \Delta y + \Delta x \Delta z + \Delta y \Delta z$.
Here, each $\Delta \zeta$ is defined by the extend of the selected cluster along the $\zeta$-axis.
This measure will grow with increased sparsity of the selection points,
and we think it is a more accurate way of representing the amount of work required for human annotators.

\section{Experiments}
In the following, we lay out the experimental part of this work starting with the different metrics used, datasets, results and ablation studies.

\subsection{Metrics}\label{subsec:metrics}

\textbf{mIoU} The mean Intersection-over-Union (mIoU) metric is commonly used to quantify the accuracy of semantic segmentation masks. It provides an average overlap between the predicted and ground-truth point labels.

\textbf{mIoU@90} The mIoU@90 is used in the AL community as the target performance.
It is $90\,\%$ of the mIoU performance that can be achieved using supervised training on the whole dataset.

\textbf{Area} We propose an alternative measure to approximate the annotation effort of a queried region. It is defined as the sum over the cluster areas as described in \cref{sec:measuring_effort}.

\subsection{Datasets}

\begin{table}[t!]
    \centering
    \begin{tabular}{lrcc}
        \toprule
        Dataset & \multicolumn{1}{c}{Points} & Semantics & Classes \\
        \midrule
        S3DIS &  273,546,486 & $100\,\%$ & 13 \\
        Toronto-3D & 78,320,210 & $100\,\%$ & 9 \\
        Freiburg  & 57,995,691,249 & $\sim 0.15\,\%$ & 13 \\
        \bottomrule
    \end{tabular}
    \caption{Overview of the datasets.
    }
    \label{tab:overview_datasets}
\end{table}

We use the Stanford 3D Indoor Scene Dataset (S3DIS)~\cite{armeni_cvpr16} and Toronto-3D~\cite{tan2020toronto3d} to evaluate our approach.
Additionally, we show the applicability of our approach on a private dataset from the city of Freiburg, Germany.
All datasets include semantic labels and colorized points.
We utilize the latter together with the XYZ locations for classification.
\cref{tab:overview_datasets} summarizes the datasets.

\textbf{S3DIS} dataset~\cite{armeni_cvpr16}, is a large indoor point cloud dataset. It is divided into six large areas and has a total of 271 rooms. For each room, a dense point cloud with color and position information is provided.
We use the 'Area5' validation set for all our performance evaluations.

\textbf{Toronto-3D} dataset~\cite{tan2020toronto3d}, is a large-scale urban outdoor point cloud dataset from Toronto, Canada, covering 1 km of road.

\textbf{Freiburg} dataset~\cite{amtsblatt_freiburg}, is a private dataset from the city of Freiburg. The LiDAR information is accompanied by RGB and intensity information.
It covers about 78\,km$^2$ with a spatial resolution in the centimeter range.
This dataset does not come with semantic labels, instead, we manually annotated a small fraction, amounting to about 42,500\,m$^2$.
The intended usage of the data for environmental climate modeling~\cite{BriegelModelling} motivates our set classes: building, wall, car, cobblestone surface, street, leaf tree, fir tree, grass, open soil, bush, hedge, vegetation, and unknown.
The annotation took about 85 working hours plus
additional time for quality assurance, including review and correction.
This illustrates the importance of AL in reducing the amount of manual labeling.
For this dataset, the percentages of labeled datta in the following sections are given with respect to the annotated subset reported in \cref{tab:overview_datasets}.

\subsection{Network Architecture, Hyperparameters and Codebase}
We evaluate all AL methods and region separation and selection strategies on two different network architectures: SPVCNN \cite{tangContrastiveBoundaryLearning2022} and the Minkunet \cite{choy4DSpatioTemporalConvNets2019}.
For comparability, we extended the codebase of~\cite{wu_redal_2021}.
Similar to them we use the Adam optimizer with a learning rate of $0.001$.
We use a batch size per GPU of $4$ and an ensemble size of $N=4$ (where applicable).
Each model was trained on a single NVIDIA RTX A6000 GPU.


\subsection{Results}\label{subsec:results}

\begin{figure*}[t!]
    \centering
     \begin{subfigure}[t]{.48\textwidth}
      \includegraphics[width=\linewidth]{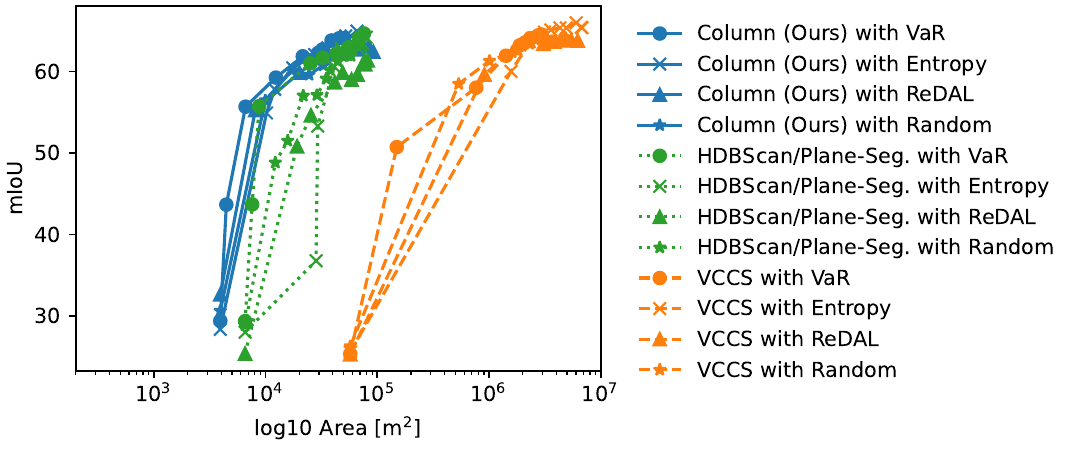}
      \caption{Outdoor: Toronto with Minkunet}
      \label{fig:toronto_iou_plot}
    \end{subfigure}%
     \begin{subfigure}[t]{.48\textwidth}
      \includegraphics[width=\linewidth]{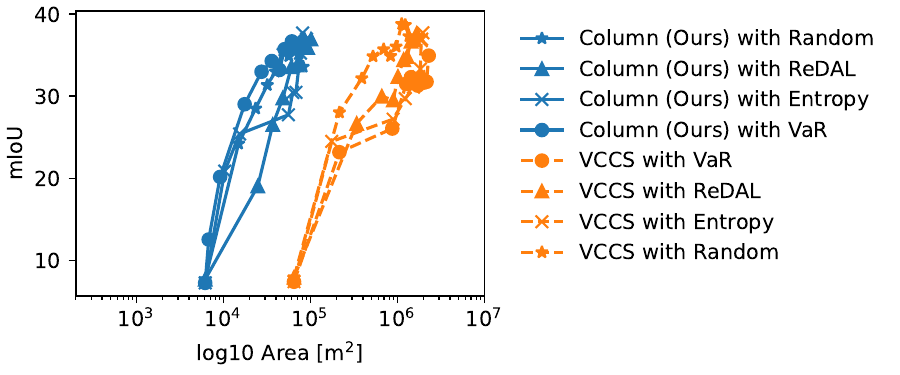}
      \caption{Outdoor: Freiburg with Minkunet}
      \label{fig:freiburg_iou_plot}
    \end{subfigure}%
    \hspace{\fill}
     \begin{subfigure}[t]{.48\textwidth}
      \includegraphics[width=\linewidth]{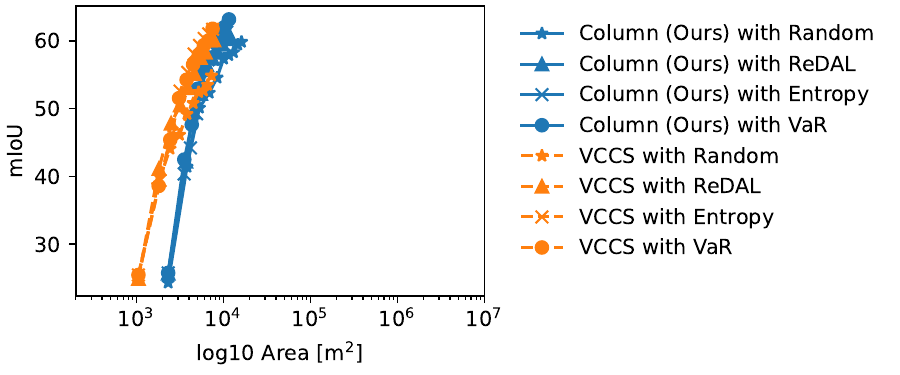}
      \caption{Indoor: S3DIS with Minkunet}
      \label{fig:s3dis_iou_plot}
    \end{subfigure}%
    \caption{Performance as a function of the annotated area for all datasets. The blue lines correspond to our proposed column separation, the orange line correspond to the region separation with VCCS and the green line correspond the the HDBScan-based separation.}
    \label{fig:iou_vs_area}
    \vspace{-.5cm}
\end{figure*}

\subsubsection{Performance with respect to covered area}

We first present our results in terms of area which is required to be labeled, which we believe is a more representative estimate of the labelling effort.
It is apparent in \cref{fig:toronto_iou_plot} and \cref{fig:freiburg_iou_plot} that our column-based separation requires way less annotated area (more than an order of magnitude) compared to  VCCS region separation, and still about a factor of two less area than HDBScan, for a similar mIoU.


For the indoor dataset S3DIS both region separation techniques show a comparable performance as function of the annotated area, as shown in \cref{fig:s3dis_iou_plot}. In this setting the VCCS based approaches are slightly better.
We suspect that this is mostly caused by the fact that the columns always include floor and ceiling.
For that reason, in indoor environments the annotated area is inflated for our approach, which could easily be disregarded by a human annotator by not considering the ceiling.
Despite this inflation, we can see that after the first AL cycles both approaches are competitive in terms of required annotated area.

\begin{figure*}[t!]
    \centering
     \begin{subfigure}[t]{.3\textwidth}
      \includegraphics[width=\linewidth]{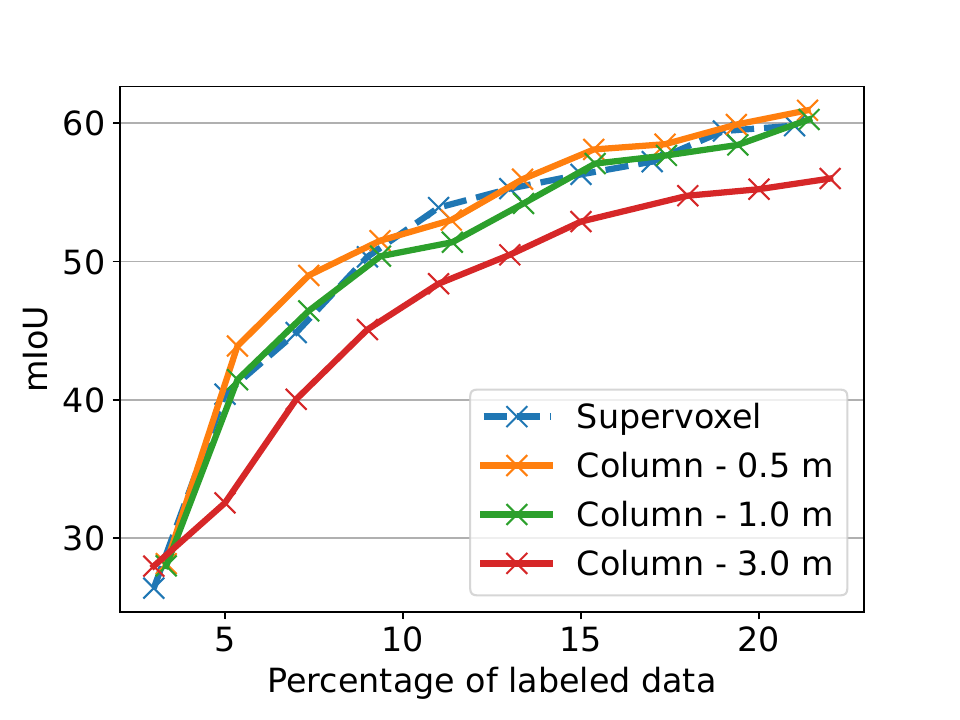}
      \caption{S3DIS with SPVCNN}
      \label{fig:s3dis_spatial}
    \end{subfigure}%
     \begin{subfigure}[t]{.3\textwidth}
      \includegraphics[width=\linewidth]{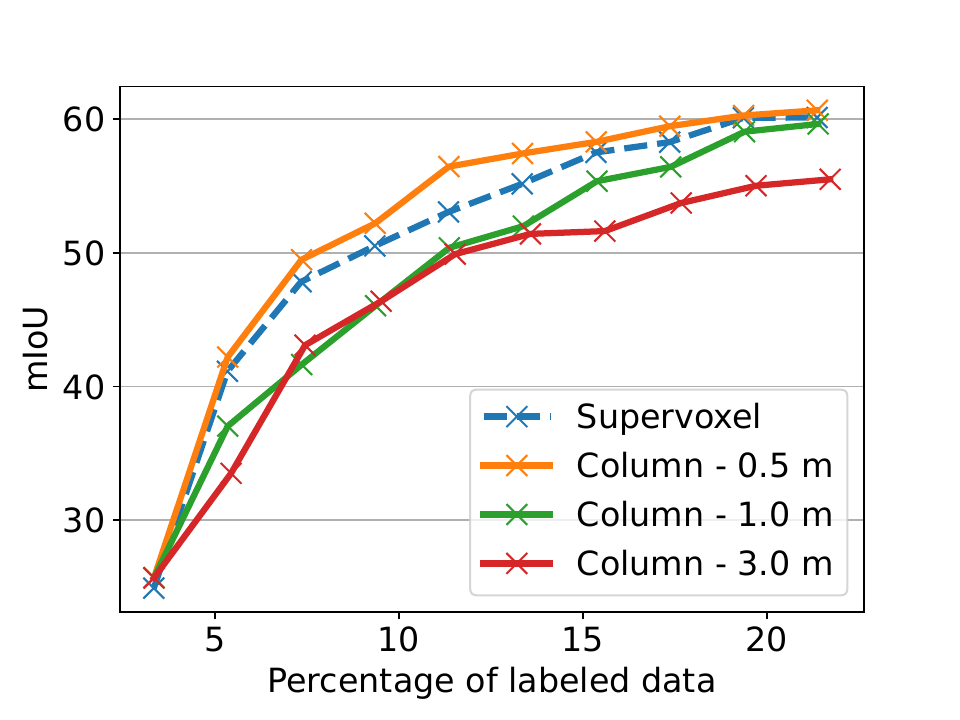}
      \caption{S3DIS with Minkunet}
      \label{fig:s3dis_spatial_minkunet}
    \end{subfigure}%
    \hspace{\fill}%
    \begin{subfigure}[t]{.3\textwidth}
      \includegraphics[width=\linewidth]{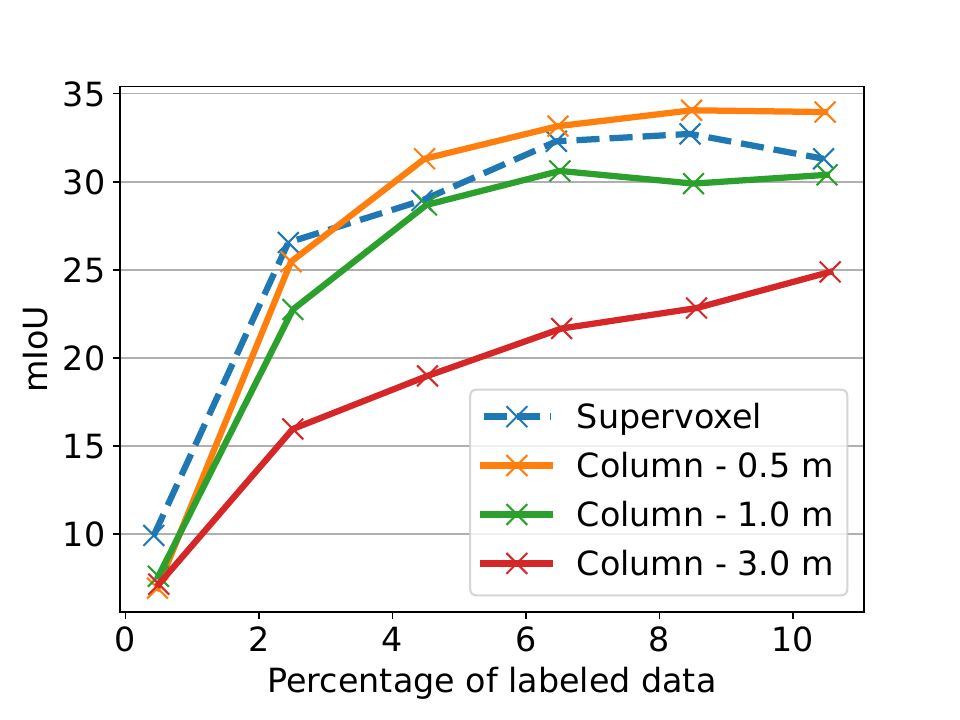}
      \caption{Freiburg with SPVCNN}
      \label{fig:freiburg_spatial}
    \end{subfigure}%
    \begin{subfigure}[t]{.3\textwidth}
      \includegraphics[width=\linewidth]{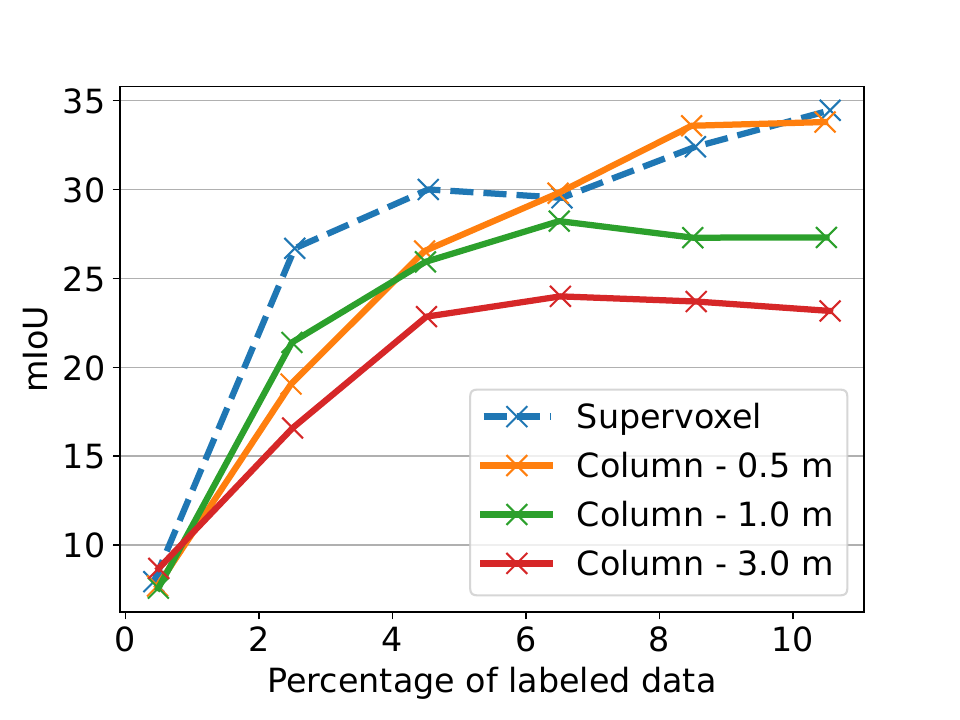}
      \caption{Freiburg with Minkunet}
      \label{fig:freiburg_spatial_minkunet}
    \end{subfigure}%
    \caption{Performance for the region separation methods
    supervoxels, created with VCCS, and columns with edge lengths of $0.5$, $1.0$, and $3.0$ meters with ReDAL as region selection algorithm.
    We show results for the S3DIS and Freiburg datasets with SPVCNN or Minkunet as segmentation models.
    }
    \label{fig:spatial_complexity}
    \vspace{-.5cm}
\end{figure*}

\begin{figure*}[ht]
    \centering
     \begin{subfigure}{.3\textwidth}
      \centering
      \includegraphics[width=\linewidth]{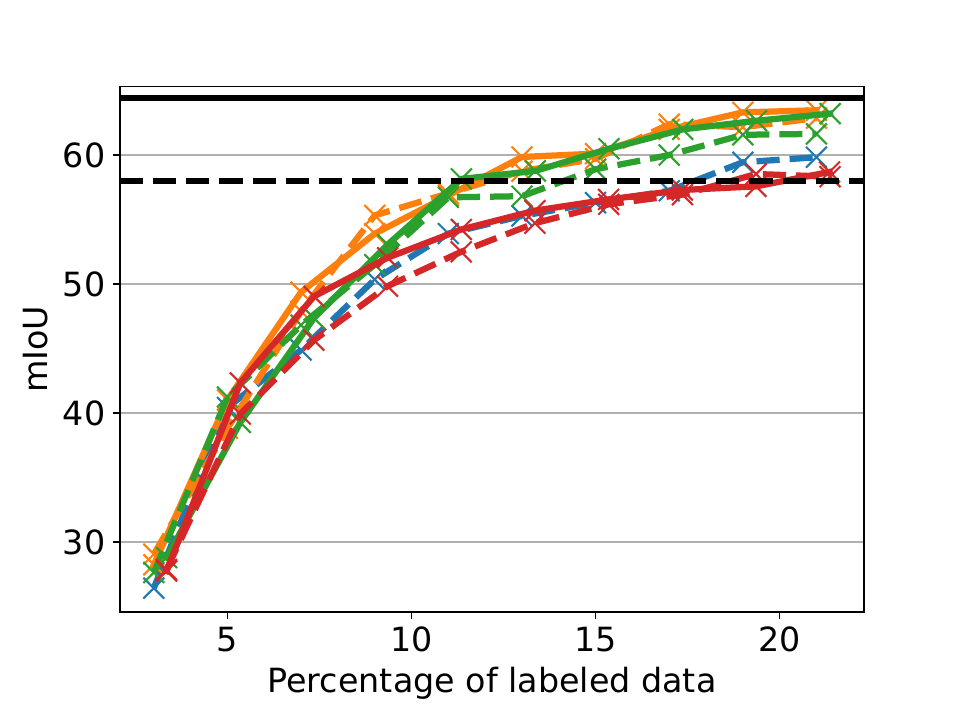}
      \caption{S3DIS - SPVCNN}
      \label{fig:s3dis_utility}
    \end{subfigure}%
    \begin{subfigure}{.3\textwidth}
      \centering
      \includegraphics[width=\linewidth]{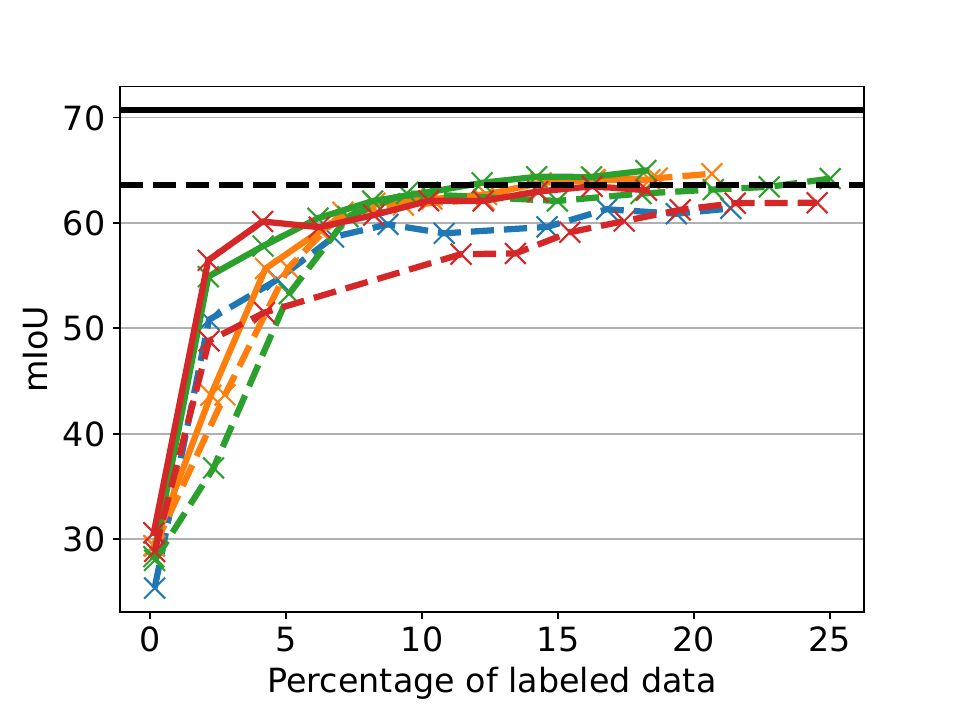}
      \caption{Toronto-3D - SPVCNN}
      \label{fig:toronto3d_spvcnn_utility}
    \end{subfigure}%
    \begin{subfigure}{.3\textwidth}
      \centering
      \includegraphics[width=\linewidth]{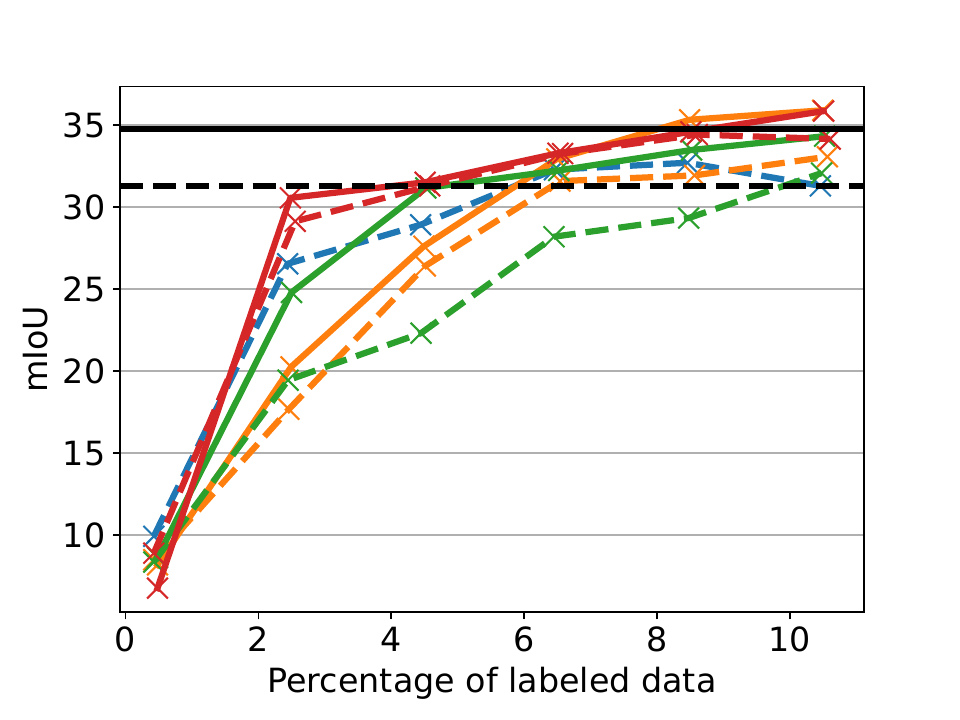}
      \caption{Freiburg - SPVCNN}
      \label{fig:freiburg_utility}
    \end{subfigure}\hspace{\fill}%
    
     \begin{subfigure}{.3\textwidth}
      \centering
      \includegraphics[width=\linewidth]{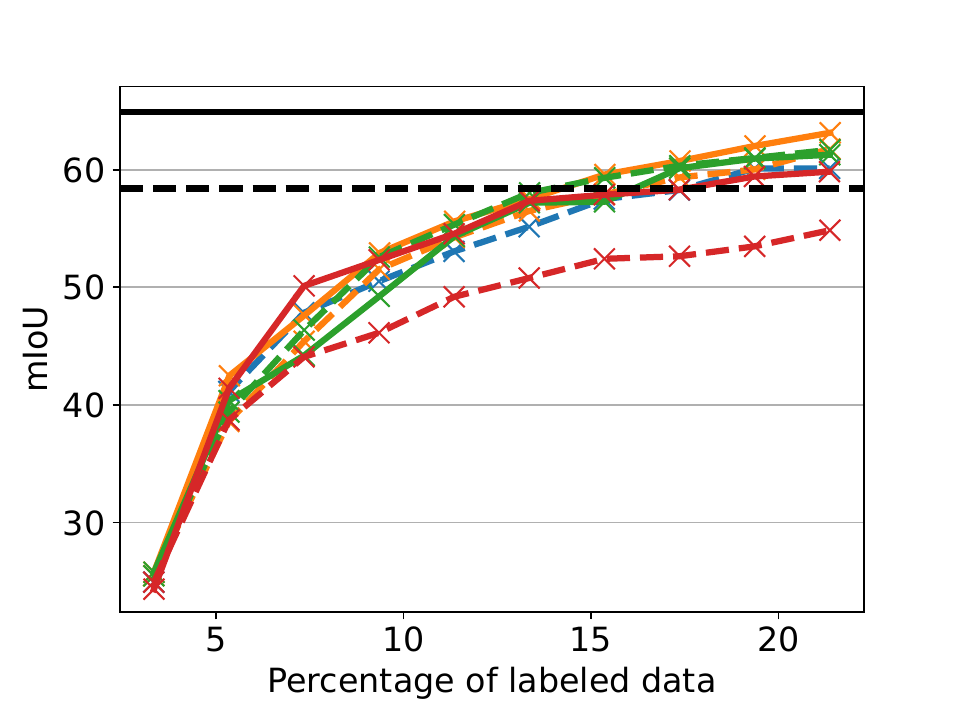}
      \caption{S3DIS - Minkunet}
      \label{fig:s3dis_minkunet_utility}
    \end{subfigure}
    \begin{subfigure}{.3\textwidth}
      \centering
      \includegraphics[width=\linewidth]{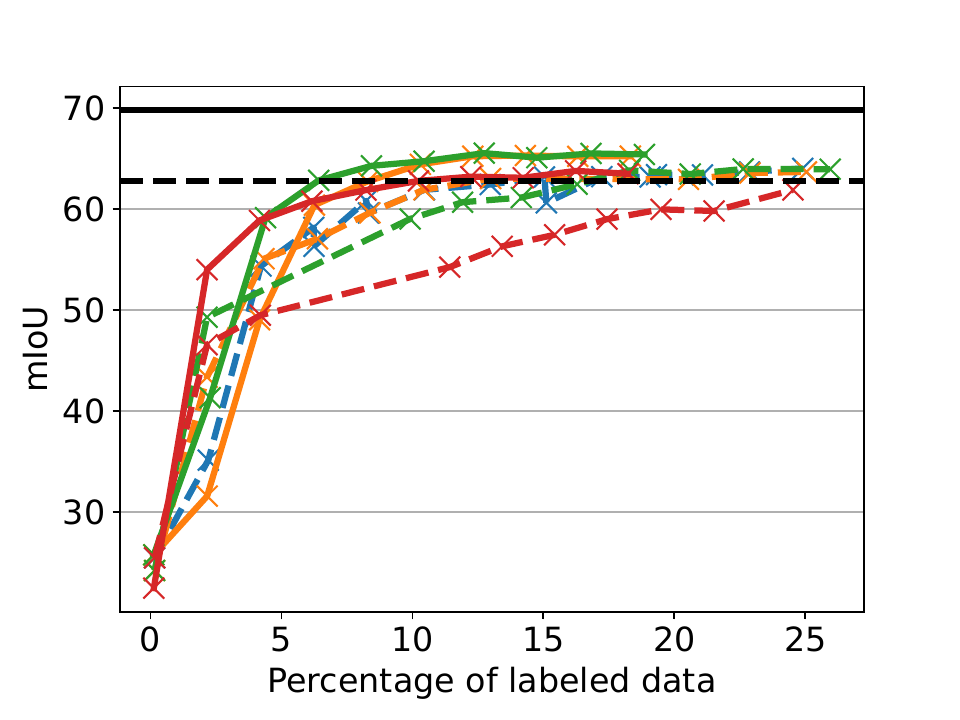}
      \caption{Toronto-3D - Minkunet}
      \label{fig:toronto3d_minkunet_utility}
    \end{subfigure}
    \begin{subfigure}{.3\textwidth}
      \centering
      \includegraphics[width=\linewidth]{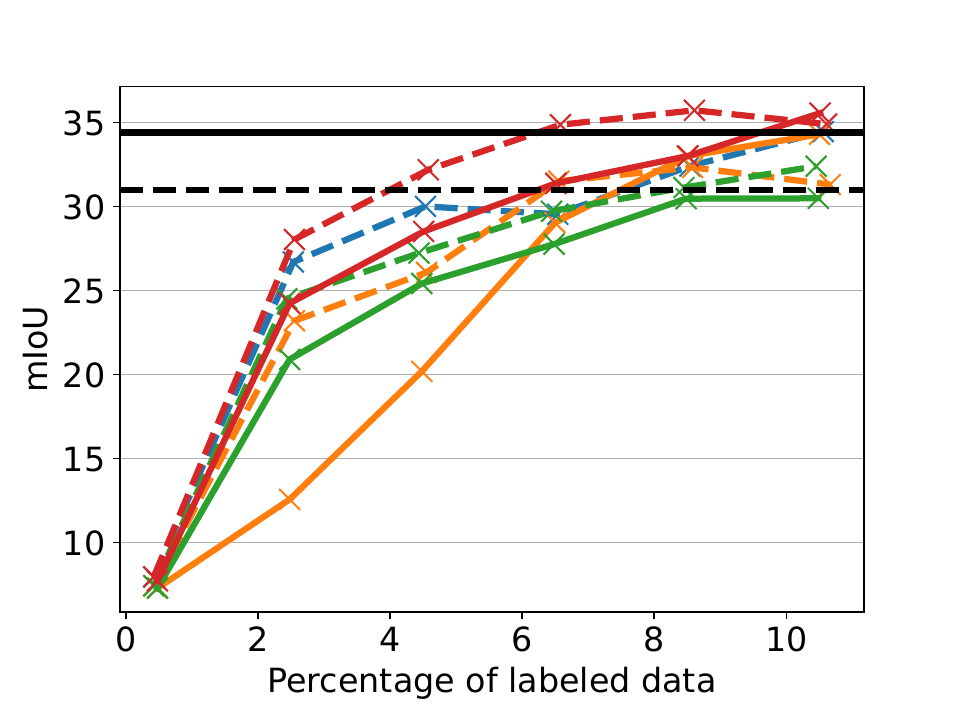}
      \caption{Freiburg - Minkunet}
      \label{fig:freiburg_minkunet_utility}
    \end{subfigure}\hspace{\fill}%
    
    \begin{subfigure}{\textwidth}
      \centering
      \includegraphics[width=0.7\linewidth]{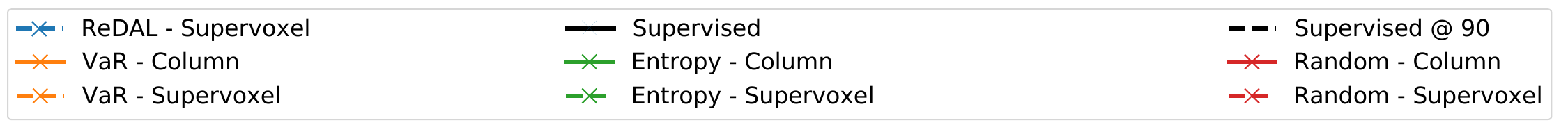}
    \end{subfigure}\hspace{\fill}%
    \caption{
    Performance for various region separation
    (supervoxels and columns of size $0.5$\,m) and selection methods
    (ReDAL, VaR, ensemble entropy, and random).
    We show results for the S3DIS, Toronto, and Freiburg datasets with SPVCNN or Minkunet as segmentation models.
    We also report the performance for supervised learning on all labeled data, as well as $90\,\%$ of that value (mIoU@$90\,\%$).
    }
    \label{fig:utility_complexity}
    \vspace{-.5cm}
\end{figure*}

In order to make our analysis comparable to other works in the community of AL, we will present the forthcoming results in terms of the percentage of labeled data.

\subsubsection{Region separation}
As first component, we evaluate the region separation mechanism. For a fair comparison, we use the ReDAL region selection but vary only the method by which the point cloud is partitioned into regions. The complete results are shown in \cref{fig:spatial_complexity}.
Here, we use the ReDAL region selection policy for comparability and only change how we split the point cloud into regions.
While the results for the S3DIS dataset are fairly consistent between the two network architectures, indicating low variance, there is more variation found in the Freiburg results.
Hence, we will focus on the S3DIS (and Toronto) datasets for our interpretation, while the Freiburg results are deemed for validation.

We observe that columns with an edge length of $0.5$~meters are competitive with the supervoxels used by ReDAL.
These results show that the performance of ReDAL is not bound to the burden of supervoxel computation.
Competitive or even better performance can be achieved by using our column-based approach.
The granularity of the regions seems to be an important factor in the performance, since with a larger edge length of $3.0$~meters the selected columns are not competitive anymore.

\subsubsection{Region selection} 
As second component, we evaluate the region selection mechanism. In this part of the evaluation, we fix the region separation mechanism to supervoxel or columns with an edge length of $0.5$ meters as they performed best with the ReDAL algorithm. We compare our ensemble-based entropy and VaR metrics, against ReDAL and the random policy. The complete results are shown in \cref{fig:utility_complexity}.


For the S3DIS dataset (\cref{fig:s3dis_utility}), we can see that all approaches show similar performance for the range of $3-7\,\%$ labeled data.
However, in the later AL stages, the ensemble-based methods outperform ReDAL as well as the random selection policies. 
We find that the entropy-based selection of columns with SPVCNN is able to reach the 
mIoU@90 threshold with $11.3\,\%$ of labeled points.
This also outperforms SSDR-AL \cite{shao_active_2022} and other methods as shown in \cref{tab:comparison_to_active_learning_spatial_structural},
and highlights the capabilities of our proposed pipeline. Comparing our results with the original ones from ReDAL~\cite{wu_redal_2021}, we find some discrepancies.
Our version of ReDAL, using the original codebase, crosses the mIoU@90 threshold at about $19\,\%$ annotated data, while the authors report $13$ to $15\,\%$.

\begin{table}[]
\centering
\begin{tabular}{ll|cc}
\toprule
 &  &  \multicolumn{2}{c}{Points} \\
In & Methods &  SPVCNN~\cite{tangSearchingEfficient3D2020} & Minkunet~\cite{choy4DSpatioTemporalConvNets2019}  \\
\midrule
\multirow{4}{2em}{Our} & Columns + Random  & 21.4\,\% & 19.4\,\% \\
& Columns + Entropy  & \textbf{11.4\,\%}  & 17.4\,\%\\
& Columns + VaR  & 13.0\,\% & \textbf{15.4\,\%} \\
& ReDAL~\cite{wu_redal_2021}  & 19.0\,\% & 19.3\,\% \\
\midrule
\cite{wu_redal_2021} & ReDAL~\cite{wu_redal_2021} & 13\,\% & \textbf{15\,\%} \\
\midrule
\multirow{5}{1em}{\cite{shao_active_2022}} & Random~\cite{shao_active_2022} & \multicolumn{2}{c}{40.9\,\%} \\
& Entropy~\cite{joshi2009multi} & \multicolumn{2}{c}{46.7\,\%} \\
& BvSB~\cite{joshi2009multi} & \multicolumn{2}{c}{43.0\,\%} \\
& ClassBal~\cite{cai2021revisiting} & \multicolumn{2}{c}{13.3\,\%} \\
& SSDR-AL~\cite{shao_active_2022} & \multicolumn{2}{c}{\textbf{11.7\,\%}} \\ 
\bottomrule
\end{tabular}
\caption{
Amount of points required for $90\,\%$ of the supervised training performance.
Our results are compared to the ones reported in~\cite{wu_redal_2021} and~\cite{shao_active_2022}.
The latter uses the RandLA-Net~\cite{hu2020randla} segmentation model.
All results are evaluated on the S3DIS dataset.
}
\label{tab:comparison_to_active_learning_spatial_structural}
    \vspace{-.8cm}
\end{table}

This might be attributed to the variance in the results, e.g. from random seeds. 
The mIoU@90 is particularly sensitive to this effect if the performance saturates around this value,
which we observe on the Toronto-3D dataset.
On Toronto-3D, we estimate the standard deviation from 5 random seeds on the mIoU to amount to
about $\pm 1.4\,\%$ in shoulder region of the curves (2-4\,\% labeled points) and to
about $\pm 0.5\,\%$ in the saturation region ($\ge 10\,\%$ labeled points). Hence, the mIoU@90 value differs by up to $\pm 3\,\%$ of labeled points. These findings sit well with the observations in Samet \emph{et al.} \cite{caronEmergingPropertiesSelfSupervised2021} which investigate the differences in seed selection for AL.


On the Toronto-3D data with SPVCNN, the column-based region separation with Entropy and VaR reaches the mIoU@90 threshold with $12\,\%$ and $14\,\%$ of annotated data, respectively.
The supervoxel-based counterparts require $16\,\%$ and $25\,\%$, respectively, while
random selection and ReDAL are unable to reach the threshold within ten AL cycles.
The results with Minkunet show a similar overall picture.


In regards to the Freiburg data, though large variances in the results are observed,
it is evident that the random policy outperforms the informed region selection methods.
We attribute this to the small overall fraction of labels in the data being more widespread, covering a relatively large area.
This is in contrast to Toronto-3D, which focuses on a single street.
Therefore, we conclude that the label set remains noisy, which also explains the network's ability to outperform the fully supervised baseline on a smaller data budget.

It should be noted that, across most datasets and segmentation models, the random selection policy also performs very well.
At first glance, this is in contradiction to some works that report very poor performances of random selection policies~\cite{wu_redal_2021,shao_active_2022}.
However, often these poor scores stem from random point-selection policies instead of random region-selection policies.

In summary, our results demonstrate that our easy-to-implement AL pipeline using spatial columns as a region separation mechanism and ensemble-based region selection policies are competitive with, or better than, state-of-the-art approaches.

\subsection{Ablation}\label{subsec:ablation}

In the following, we show that our approach is computationally cheaper and further analyze which augmentations are the most important during training. This is especially important when scaling to very large datasets. 

\begin{figure}[t!]
    \centering
    \includegraphics[width=0.7\linewidth]{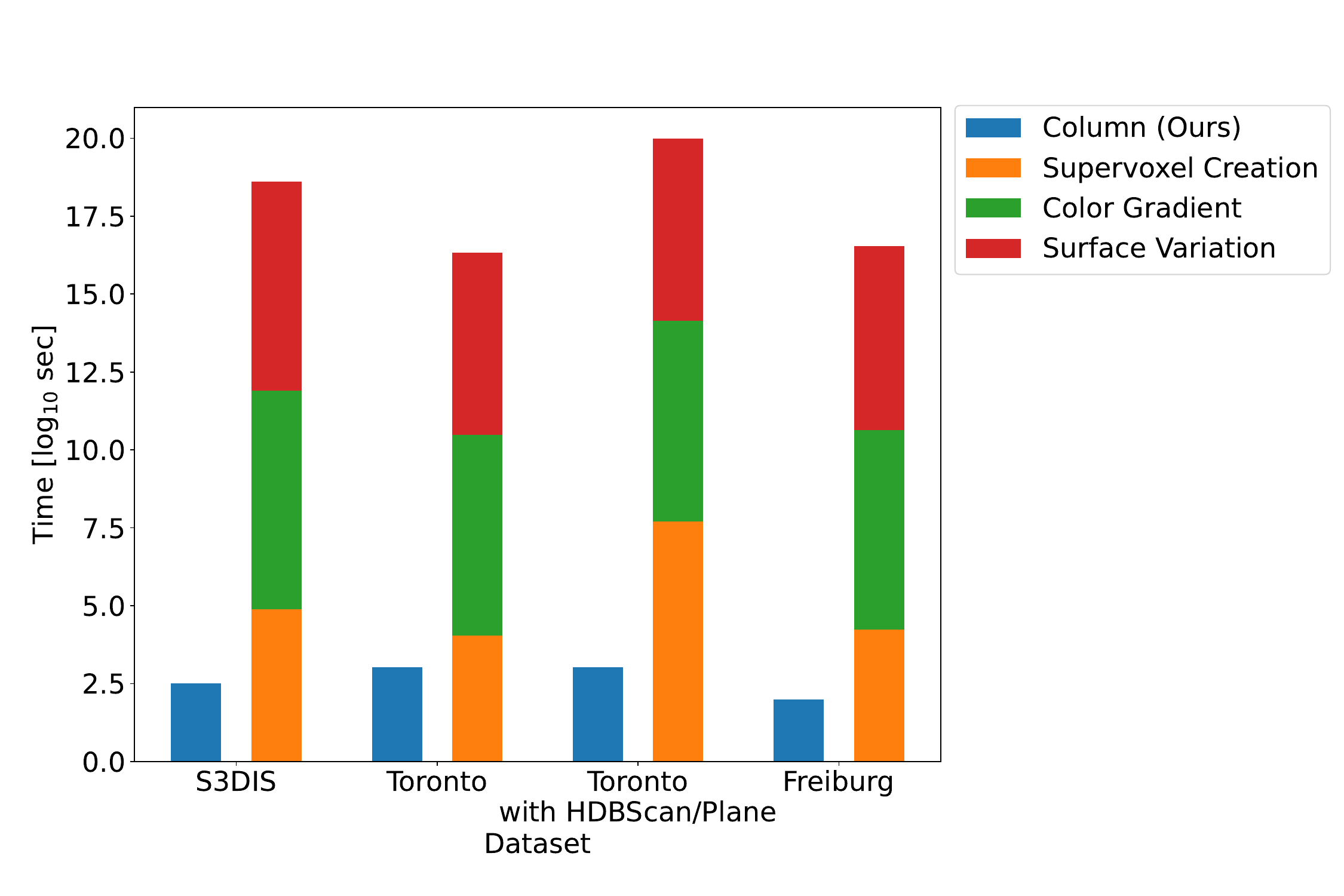}
    \caption{Comparison of preprocessing time required for our method against the preprocessing time required by ReDAL. For a fair comparison, we use the same framework as ReDAL and precompute the column-based region separation beforehand. Note, that we report log sec as duration of individual steps for a better visual comparison. Time measurement done on a machine with Ryzen 9 pro 7945 and 64 GB RAM.}
    \label{fig:preprocessing_times}
\end{figure}


\subsubsection{Preprocessing time} In \cref{fig:preprocessing_times} we compare the preprocessing time required for our column-based region separation technique and the more heuristic based method of ReDAL. In comparison all preprocessing steps individually take longer than the computation of the column-based separation. For a pointcloud with $50 \times 50 \text{m}$ the computation of column separation takes $\sim 0.3~\text{sec}$, which would also allow for online computation, but not used for a fair comparison of the methods. 
Limiting the pre-processing to the preprocessing of spatial columns reduces the pre-processing time on S3DIS to $0.59\%$ of the otherwise required time. On Toronto-3D to $1.9\%$ or $0.64\%$ with HDBScan and on the annotated portion of the Freiburg data to $0.70\%$.

\subsubsection{Data Augmentation} is one of the key techniques to diversify the training data in order to achieve better training results. However, it is often very domain-dependent on which data augmentation techniques perform the best. As methods like RandAugment \cite{cubuk2020randaugment} or TrivialAugment \cite{Muller_2021_ICCV} have shown, it is often the mixture of different augmentations that works the best. To improve the AL cycle in terms of performance and required time, we investigate the influence of the individual augmentation methods used in ReDAL \cite{wu_redal_2021}. \\

\cref{tab:data_augmentation} shows the influence of the different data augmentation methods for $2\,\%$, $10\,\%$, and $100\,\%$ percent of data files used. Note, that the sub-$100\,\%$ data points are randomly sampled and not selected by any method. From the results, one can observe that generating novel views of the data through rotation is by far the most important augmentation method. In the lower-data regime, it outperforms the combination of all other data augmentation schemes. However, when training with all data, training with all augmentations is still the best.

\begin{table}[]
    \centering
    \begin{tabular}{c|cc|cc|cc}
        \toprule
        & \multicolumn{2}{c|}{2\,\% labels} & \multicolumn{2}{c|}{10\,\% labels} & \multicolumn{2}{c}{100\,\% labels} \\
          \thead{S/R/E/C} & mIoU & \thead{Time} & mIoU & \thead{Time} & mIoU & \thead{Time}  \\
         \midrule
            $\times$  $\times$  $\times$  $\times$ &  $23.32$ &  $16.02$ &  $32.77$  & 21.67 &  $43.68$  & 159.00 \\
            $\checkmark$  $\times$  $\times$   $\times$ &  $24.48$ & $\times0.95$ &  $38.03$ & $\times1.05$ & $48.71$ & $\times1.38$  \\ 
            $\times$  $\checkmark$  $\times$   $\times$ &  $\textbf{27.55}$ & $\times1.16$  &  $\textbf{43.23}$ & $\times1.48$   & $58.79$  & $\times1.76$ \\ 
            $\times$  $\times$  $\checkmark$   $\times$ &  $21.96$ & $\times1.94$ &  $36.83$ & $\times2.56$ & $45.95$ & $\times3.33$ \\ 
            $\times$  $\times$  $\times$   $\checkmark$ &  $22.86$ & $\times1.03$ &  $32.77$ & $\times1.20$ & $44.22$ & $\times1.60$   \\ 
            $\checkmark$  $\checkmark$  $\checkmark$   $\checkmark$ &  $26.05$ & $\times2.37$ &  $41.76$ & $\times3.34$ & $\textbf{61.13}$ & $\times5.92$  \\ 
         \bottomrule
    \end{tabular}
    \caption{Performance on the S3DIS dataset using 
    data augmentation techniques:
    scale (S), rotation (R), elastic (E), and chromatic (C).
    The training time is given in minutes, or as scale factor in proportion to the first row.
    }
    \label{tab:data_augmentation}
    \vspace{-.8cm}
\end{table}

The results show that rotation augmentation is by far the most important augmentation for the performance of the network. For the low-label regime, with $2\,\%$ or $10\,\%$ of scenes sampled, it is often beneficial to only use the rotation augmentation both in terms of performance and used training time. When training with all data, it is beneficial to train with all augmentations. However, it is also the slowest training. Hence, we chose to deactivate the elastic distortion during the AL cycles of our method but to enable all other augmentations.

\section{Conclusions}

In this paper, we presented a novel active learning pipeline in the context of point cloud segmentation
that provides comparable or better performance than state-of-the-art results,
while employing easy-to-implement methods.
We evaluated our approach in the context of large-scale urban
point clouds, with classes directed at forecasting extreme
weather events, but also on the common S3DIS indoor
dataset.

In terms of region separation, we proposed to divide the
point cloud into a 2D grid of columns. Columns can be
easily optimized due to having a single parameter (the edge length) and efficiently stored without the need for point indices.
Furthermore, columns are robust under domain changes, while we observed that the more involved VCCS method can fail. Concerning the region selection step, we propose to use common ensemble uncertainty metrics, with better or equally good results as more involved hybrid approaches. This reduces the number of cumbersome preprocessing steps.

Additionally, we proposed a novel metric to determine the annotation costs of different active learning approaches.
This metric not only takes into account the number of points to be annotated but also the area that needs to be considered by the human annotator during the labeling process.
As a result, we estimate that our active learning approach for point cloud data requires less work from human annotators.
 
Despite these encouraging results, there are several aspects that warrant future research. First, one could investigate whether foundation models can be employed to replace the human annotator. Second, one could utilize a location-dependent adjustment of the grid resolution.




\subsubsection*{Acknowledgements} We acknowledge the city of Freiburg for providing us with the colored LiDAR data from the city of Freiburg.
This research was funded by the German Federal Ministry for the Environment, Nature Conservation and Nuclear Safety (BMU) on the basis of a resolution of the German Bundestag as part of the ‘KI-Leuchtturm’ project ‘Intelligence for Cities’ (I4C).

{\footnotesize
\bibliographystyle{IEEEtran}
\bibliography{references.bib,AktivesLernen.bib}
}

\end{document}